\documentclass[letterpaper]{article} 
\usepackage[draft]{aaai2026}  
\usepackage{times}  
\usepackage{helvet}  
\usepackage{courier}  
\usepackage[hyphens]{url}  
\usepackage{graphicx} 
\usepackage{natbib}  
\usepackage{caption} 
\usepackage{algorithm}
\usepackage{algpseudocode}
\usepackage{xspace}
\usepackage{multirow}
\usepackage[table]{xcolor}
\usepackage{amsmath}
\newcommand{\our}{{AffordDex}\xspace}
\usepackage{booktabs}   
\usepackage{amssymb}   
\usepackage{url}
\usepackage[colorlinks=true, allcolors=black]{hyperref}

\usepackage{cleveref}

\usepackage{newfloat}
\usepackage{listings}
\usepackage{booktabs}
\DeclareCaptionStyle{ruled}{labelfont=normalfont,labelsep=colon,strut=off} 
\floatstyle{ruled}
\newfloat{listing}{tb}{lst}{}
\floatname{listing}{Listing}
\title{Towards Affordance-Aware Robotic Dexterous Grasping with Human-like Priors}

\author{
    Haoyu Zhao *\textsuperscript{\rm 1, 2},
    Linghao Zhuang *\textsuperscript{\rm 1},
    Xingyue Zhao *\textsuperscript{\rm 2},
    Cheng Zeng\textsuperscript{\rm 5},\\
    Haoran Xu\textsuperscript{\rm 4},
    Yuming Jiang\textsuperscript{\rm 2},
    Jun Cen\textsuperscript{\rm 2, 3, 4},
    Kexiang Wang\textsuperscript{\rm 2},
    Jiayan Guo\textsuperscript{\rm 2},\\
    Siteng Huang\textdagger \textsuperscript{\rm 2, 3, 4},
    Xin Li\textsuperscript{\rm 2, 3},
    Deli Zhao\textsuperscript{\rm 2, 3},
    Hua Zou\textdagger \textsuperscript{\rm 1}
}
\affiliations{
    \textsuperscript{\rm 1} Wuhan University
    \textsuperscript{\rm 2} DAMO Academy, Alibaba Group
    \textsuperscript{\rm 3} Hupan Lab
    \textsuperscript{\rm 4} Zhejiang University
    \textsuperscript{\rm 5} Tsinghua University
}

\usepackage{bibentry}

\begin{document}

\maketitle

\maketitle
{
\renewcommand{\thefootnote}{\fnsymbol{footnote}}
\footnotetext{* Equal contributions.}
\footnotetext{\textdagger Corresponding Author}
}

\begin{abstract}
A dexterous hand capable of generalizable grasping objects is fundamental for the development of general-purpose embodied AI. However, previous methods focus narrowly on low-level grasp stability metrics, neglecting affordance-aware positioning and human-like poses which are crucial for downstream manipulation. 
To address these limitations, we propose \textbf{\our}, a novel framework with two-stage training that learns a universal grasping policy with an inherent understanding of both motion priors and object affordances. 
In the first stage, a trajectory imitator is pre-trained on a large corpus of human hand motions to instill a strong prior for natural movement.
In the second stage, a residual module is trained to adapt these general human-like motions to specific object instances.
This refinement is critically guided by two components: our Negative Affordance-aware Segmentation (NAA) module, which identifies functionally inappropriate contact regions, and a privileged teacher-student distillation process that ensures the final vision-based policy is highly successful.
Extensive experiments demonstrate that \our not only achieves universal dexterous grasping but also remains remarkably human-like in posture and functionally appropriate in contact location.
As a result, \our significantly outperforms state-of-the-art baselines across seen objects, unseen instances, and even entirely novel categories.
Here is our project page: \hyperref[https://afforddex.github.io/]{https://afforddex.github.io/}.
\end{abstract}


\begin{figure}[!t]
  \centering
  \includegraphics[width=0.9\linewidth]{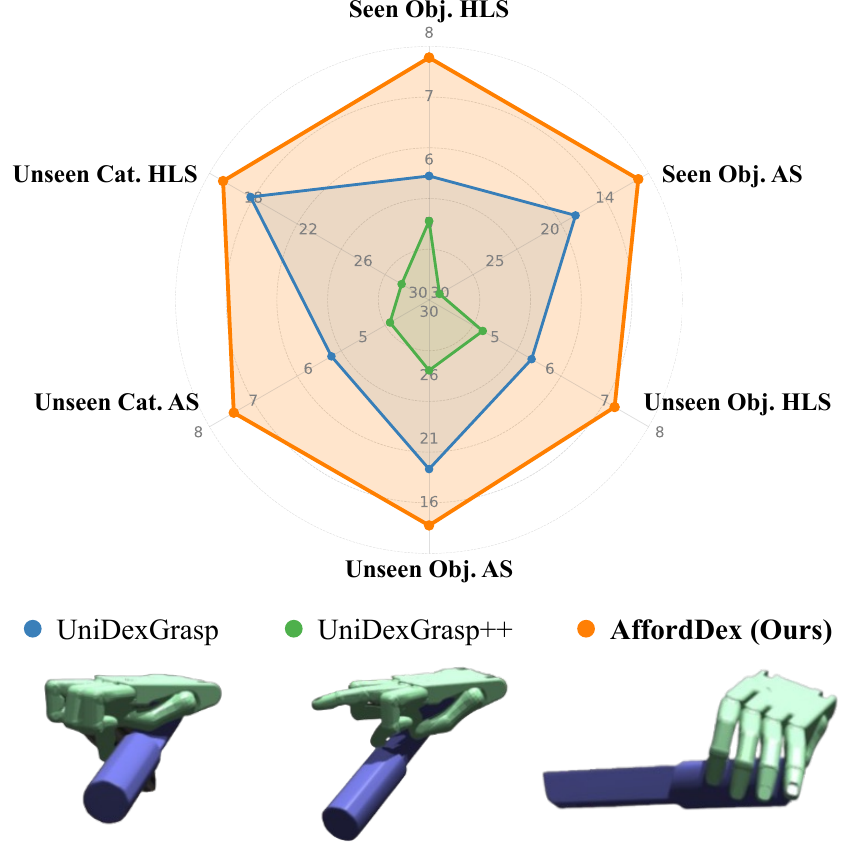}
    \vspace{-10pt}
  \caption{Performance comparision among UniDexGrasp~\cite{xu2023unidexgrasp}, UniDexGrasp++~\cite{wan2023unidexgrasp++}, and our \our, on the vision-based setting. we report human-likeness score (HLS) and affordance score (AS) across seen objects, unseen objects, and unseen categories. We also present a qualitative comparison, where \our performs natural and safe grasping by avoiding the blade.}
  \vspace{-2mm}
  \label{fig:teaser}
\end{figure}

\section{Introduction}
Dexterous grasping, as a foundational capability for robotic manipulation, has garnered significant attention from both academia and industry~\cite{zhao2024sg,zhao2025high}. Compared to simpler end-effectors (e.g., parallel jaws, vacuum grippers), five-fingered dexterous hands closely resemble human hand structure, providing substantially enhanced flexibility, precision, and task adaptability~\cite{zhong2025dexgrasp}. Furthermore, anthropomorphic robots expedite the collection of rich human demonstration data via teleoperation~\cite{li2025teleopbench}. 
Consequently, this synergy has fueled rapid progress, with recent algorithms achieving high success rates in generalizing grasps to novel objects~\cite{fang2022transcg,fang2020graspnet,gou2021rgb,wang2021graspness,xu2023unidexgrasp,wan2023unidexgrasp++}.

Due to the high degrees of freedom (DOFs) of dexterous hands, traditional motion planning-based methods~\cite{andrews2013goal,bai2014dexterous} struggle to handle such complex hand joint movements. Recent advancements in reinforcement learning (RL)~\cite{wan2023unidexgrasp++,mandikal2022dexvip,christen2022d,nagabandi2020deep,mandikal2021learning} have shown promising results in complex dexterous manipulation. However, the goal of grasping 
is not merely to lift an object.
It involves alignment with human intent and preparation for subsequent manipulation tasks, such as avoiding the blade of a knife or preparing to open a bottle cap.
Existing methods, while focused on low-level grasp stability metrics, largely overlook this crucial synthesis of affordance-aware positioning and human-like kinematics, limiting their utility in real-world, multi-step manipulation scenarios.

In this work, we focus on the critical aspect of safety and functional correctness by modeling \textit{negative} affordances—regions to be avoided, which provide clear, unambiguous negative constraints and thus simplify the learning problem.
We propose \textbf{\our}, a novel framework that learns a universal grasping policy that is both human-like in its motion and functionally aware of object affordances. 
We achieve this through a structured, two-stage training paradigm. In the first stage, we pre-train a base policy on a large corpus of human hand motions to instill a strong prior for natural movement.
In the second stage, a residual module is trained to adapts the general human-like motions from the pre-trained policy to specific objects.
This refinement is critically guided by our proposed Negative Affordance-aware Segmentation (NAA) module, which 
provides explicit visual-geometric constraints on functionally inappropriate contact regions.
Moreover, the training is enhanced with a teacher-student distillation framework, which leverages ground-truth state information to ensure the final vision-based policy is highly effective and robust.

As illustrated in Fig.~\ref{fig:teaser}, \our produces grasps that are not only successful but also remarkably human-like and functionally correct, such as safely grasping a knife by its handle. Extensive experiments validate that our method significantly outperforms existing approaches across benchmarks of seen objects, unseen objects, and even objects across datasets.
In summary, \our makes the following contributions:
\begin{itemize}
\item We propose \textbf{\our}, a two-stage framework that synergistically and effectively integrates human motion priors with functional affordance constraints to achieve generalizable and anthropomorphic dexterous grasping. 
\item We introduce a Negative Affordance-aware Segmentation (NAA) module that, by reformulating segmentation as a VLM-guided classification problem, provides explicit geometric constraints to prevent functionally improper grasps.
\item Extensive experiments demonstrate \our achieves SOTA success rates across multiple levels of generalization while producing grasps that are qualitatively superior in human-likeness and functional appropriateness.  
\end{itemize}


\begin{figure*}[t]
  \centering
  \includegraphics[width=0.95\linewidth]{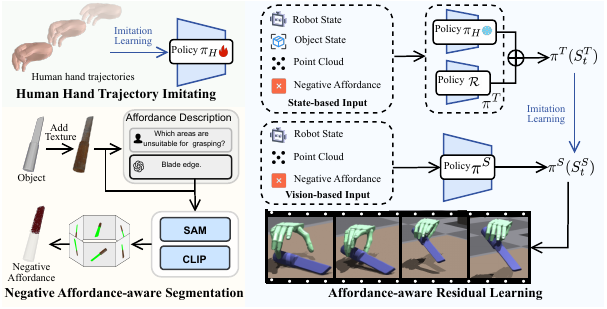}
    \vspace{-7pt}
  \caption{\textbf{Pipeline of \our}. To generate grasps with affordance-aware positioning and human-like kinematics, crucial for facilitating downstream manipulation, we propose a novel two-stage framework. The first stage establishes a strong human motion prior by training a base policy $\pi^H$, on a human motion dataset via imitation learning. This constrains the policy to a space of natural, human-like movements. Subsequently, the second stage employs reinforcement learning (RL) to refine this coarse policy $\pi^H$ for precise, functional interaction. We fine-tune $\pi^H$ with a residual module that is guided by our Negative Affordance-aware Segmentation (NAA) module, which provides explicit constraints on where not to touch the object. The entire learning pipeline is further enhanced by a teacher-student distillation framework, leveraging privileged inputs to significantly boost the final grasping performance.}
  \label{fig:pipeline}
\end{figure*}

\section{Related Work}
\subsection{Dexterous Grasping}
Robotic grasping~\cite{fang2022transcg,fang2020graspnet,gou2021rgb,wang2021graspness} has been a longstanding research, aiming to enable robots to interact with objects reliably and adaptively. 
While significant advances have been made with simple parallel-jaw grippers~\cite{fang2020graspnet,mahler2019learning}, their limited dexterity restricts adaptability to objects with intricate geometries. Dexterous, multi-fingered hands~\cite{xu2023unidexgrasp,wan2023unidexgrasp++} offer a solution but pose a severe control challenge for traditional analytical methods~\cite{bai2014dexterous,liu2021synthesizing}, motivating the shift towards learning-based approaches.

One paradigm decouples the grasping process into static grasp pose generation followed by a dynamic grasping through trajectory planning or goal-conditioned reinforcement learning (RL)~\cite{wan2023unidexgrasp++,christen2022d,wang2025unigrasptransformer}. For example, UniDexGrasp++ proposes geometry-aware curriculum learning and leverages the geometry feature for RL. However, these RL-based methods may produce physically unrealistic joint configurations. An alternative paradigm directly learns the entire grasping trajectory through expert demonstrations from humans or reinforcement learning agents~\cite{xu2023unidexgrasp,liu2024realdex,huang2023diffusion,lu2024ugg,zhang2024artigrasp,zhang2024graspxl}. 
These approaches tend to achieve more natural motions but suffer from poor generalization to novel objects due to the limited diversity of demonstrations and inherent policy constraints.
To address these failures, our \our combines strong, human-derived motion priors to ensure natural movement with affordance-based guidance to achieve robust generalization, resulting in a policy that is both natural and functionally effective across a wide range of objects.

\subsection{Affordance Prediction}
Affordance defines the action possibilities an object offers to an agent~\cite{gibson2014theory}.
In robotics, this translates to identifying object regions suitable for specific interactions, such as grasping, pushing, or lifting.
Predicting such affordances is therefore critical for advanced visual understanding and robotic manipulation, as evidenced by extensive research~\cite{li2019putting,cao2020long,corona2020ganhand,jiang2021hand,lu2025geal,shao2025great}. 
While pioneering works like GanHand~\cite{corona2020ganhand} introduced generative models for multi-object on-table grasps, and GEAL~\cite{lu2025geal} pioneered a dual-branch architecture for cross-modal (3D point cloud to 2D) representation learning, their learned affordances are often task- or category-specific. This inherent specialization limits their ability to generalize to novel objects or adapt to different downstream manipulation requirements. By contrast, humans exhibit exceptional proficiency in inferring universal affordances from visual cues~\cite{zhao2024automated}. Inspired by this capability, \our learns to infer functional affordances directly from multi-view rendered images of 3D objects, enabling a generalizable grasping policy that is not constrained to specific object categories or predefined tasks.

\section{Methodology}
To generate grasps with affordance-aware positioning and human-like kinematics, crucial for facilitating downstream manipulation, we propose a novel two-stage framework. The first stage establishes a strong human motion prior by pre-training a base policy $\pi^H$, on a large-scale human motion dataset~\cite{zhan2024oakink2} via imitation learning. This constrains the policy to a manifold of natural, human-like movements. In the second stage, we freeze the weights of $\pi^H$ and train a lightweight residual module via reinforcement learning (RL) to adapt these general motions to specific object interactions. This RL refinement stage is critically guided by two components: our Negative Affordance-aware Segmentation (NAA) module, which provides explicit constraints on where not to touch an object, and a teacher-student distillation framework that leverages privileged state information to significantly boost the final policy's performance. An overview of our method is illustrated in Fig.~\ref{fig:pipeline}.


\subsection{Human Hand Trajectory Imitating}
\label{sec:hand_imit}
In this stage, our objective is to learn a base policy $\pi^H$, that captures the kinematic priors of natural human hand motions. We formulate this task as a reinforcement learning (RL) problem where the policy $\pi^H(a_t | S_t^H)$ learns to generate dexterous hand action $a_t$ based on the current state $S_t^H$ at time $t$. To facilitate the following fine-tuning stage, the state consists of robot state $R_t$, object state $O_t$, and point cloud representation of object $P_t$, i.e., $S_t^H = \{ R_t, O_t, P_t \}$.

\medskip
\noindent
\textbf{Reward function.} 
We design a reward function $r^H$ to promote both precise imitation of human hand trajectories and the motion stability. It is composed of two terms: a finger imitation reward $r^H_{\text{finger}}$ and a smoothness reward $r^H_{\text{smooth}}$. 

The finger imitation reward $r^H_{\text{finger}}$ encourages the dexterous hand to closely track the reference finger poses from human hand dataset. Following~\cite{li2025maniptrans}, we define this reward based on the distance between the corresponding keypoints $F$ on the robot dexterous hand and the MANO hand. The reward at time $t$ is formulated as:
\begin{align}
    r^H_{\text{finger}} = \sum_{f=1}^{F} w_{f} \cdot \exp \left( -\lambda_{f} \left\| \mathbf{j}_{d, f} - \mathbf{j}_{h, f} \right\|_{2}^{2} \right),
\label{eq:eq1}
\end{align}
where $\mathbf{j}_{d, f}$ is the position of the $f$-th keypoint on the dexterous hand, $\mathbf{j}_{h, f}$ is its corresponding target position from the reference trajectory, $w_{f}$ is weight and $\lambda_{f}$ is the decay rate.

The smoothness reward $r^H_{\text{smooth}}$ encourages energy-efficient movements by penalizing excessive power consumption. This is computed as the element-wise product of joint velocities and applied torques. A detailed formulation of our reward function is available in Supp. Mat.


\subsection{Negative Affordance-aware Segmentation}
\label{sec:afford}
A significant limitation of prior work in grasp synthesis~\cite{xu2023unidexgrasp,wan2023unidexgrasp++,zhong2025dexgrasp}, is its neglect of the semantic and functional context of the interaction. A classic example is a knife: while its blade is geometrically stable for grasping, any such grasp is functionally incorrect and unsafe. To address limitation, we introduce the Negative Affordance-aware Segmentation (NAA) module to incorporate negative affordances—reasoning about which parts of an object should not be touched. The proposed NAA has the ability to operate in an open-vocabulary manner by harnessing the rich world knowledge embedded in Vision-Language Models (VLMs)~\cite{radford2021learning,achiam2023gpt}, automatically benefiting from future progress in foundation models. This ensures that the generated grasps are not only geometrically stable but also semantically coherent and task-aware.

VLMs struggle to interpret non-textured 3D meshes, as these models primarily rely on rich visual cues learned from images. To bridge this gap, we first apply procedural texturing to the raw meshes by~\cite{zhang2024texpainter}, which generates semantically plausible textures based on geometric analysis, ensuring robustness across different object shapes. Next, we render the textured object from six cardinal directions to create a multi-view image set $I$ as a holistic visual representation.
While this may not capture all concavities in highly complex objects, we found it provides a sufficient basis for affordance prediction for objects in the benchmark datasets, representing a practical trade-off between coverage and computational cost.
We then query GPT-4V~\cite{achiam2023gpt} to elicit a detailed description of the object's negative affordances. 

VLMs~\cite{radford2021learning} and Multimodal Large Language Models (MLLMs)~\cite{achiam2023gpt} excel at image-level understanding but struggle with the fine-grained spatial localization required for segmentation. To solve this, instead of asking CLIP~\cite{radford2021learning} to find ``blade part'' from the image, we turn the segmentation task into a much simpler classification task. We generate a set of precise object-part masks $M_i$ and use them as a visual prompt to let CLIP identify which mask in $M_i$ has the highest semantic similarity to the textual description ``\textit{blade part}''.

Specifically, for each image $I_i \in I$, we prompt Segment Anything Model (SAM)~\cite{kirillov2023segment} with a dense grid of points $G$ overlaid on $I_i$, which prompts SAM to perform an exhaustive segmentation, identifying all potential objects and parts. The resulting collection of masks is then refined using Non-Maximum Suppression (NMS) to eliminate duplicates, yielding a clean candidate masks set $M_i$: 
\begin{align}
M_i = \text{NMS}(\text{SAM}(I_i, G_i)).
\end{align}

\begin{figure}[t]
  \centering
  \includegraphics[width=\linewidth]{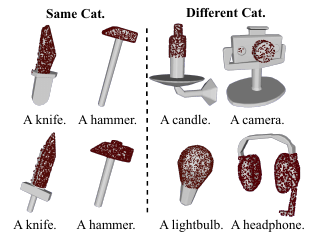}
  \vspace{-20pt}
  \caption{\textbf{Visualization} of Negative Affordances Predicted by our NAA. The point cloud, highlighted in red, represents the negative affordances identified on various objects. These points denote regions that are functionally unsafe or inappropriate for grasping, such as a knife's blade.}
  \label{fig:naa_show}
\end{figure}

For each mask $M_i^j \in M_i$, we generate a visually prompted image $I_i^j$ by blurring regions outside the mask with Gaussian filter following~\cite{yang2023fine}. The prompted image set $\{ I_i^j \}$ is then passed to CLIP along with the text query to compute a similarity score for each image-text pair. The mask with the highest similarity score is selected as the final segmentation mask. The mask is then projected into 3D space to segment the corresponding regions of the object's point cloud to get the negative affordance $N_t$, as shown in Fig.~\ref{fig:naa_show}. Our NAA is an offline, one-time process. About 160 seconds per object on an RTX 4090 is a reasonable one-time trade-off.


\begin{figure*}[t]
  \centering
  \includegraphics[width=0.95\linewidth]{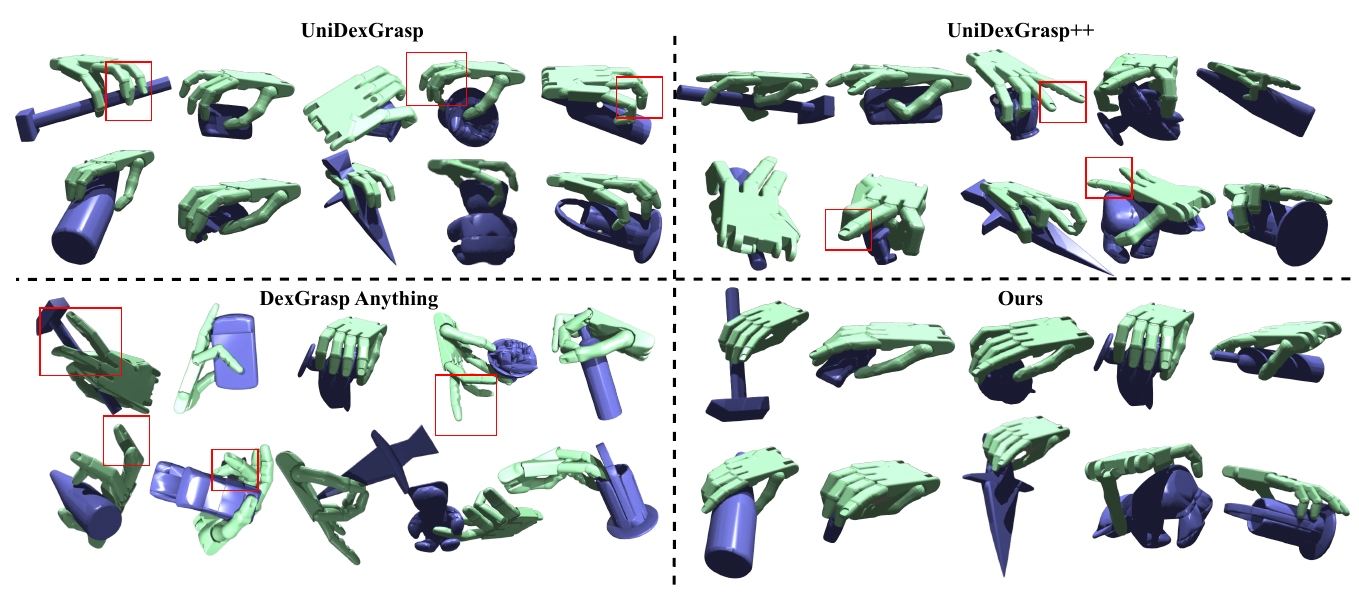}
  \vspace{-12pt}
  \caption{\textbf{Qualitative Comparison} on UniDexGrasp~\cite{xu2023unidexgrasp} and OakInk2~\cite{zhan2024oakink2}. A comparison of grasps generated by our \our with several baselines, including UniDexGrasp~\cite{xu2023unidexgrasp}, UniDexGrasp++~\cite{wan2023unidexgrasp++}, and DexGrasp Anything~\cite{zhong2025dexgrasp}.}
  \label{fig:exp}
\end{figure*} 


\subsection{Affordance-aware Residual Learning}
The negative affordance predicted from proposed NAA, we use a residual module $\mathcal{R}$ to refine the pre-trained policy $\pi^H$.
Since visual pose estimation is inherently less precise than using privileged state information, directly training an effective vision-based policy can be challenging. Therefore, we first train a state-based teacher policy $\pi^T$ which can access the ground-truth states of the environment, such as object states, to learn residual actions to refine the initial actions predicted by $\pi^H$. Once the teacher policy $\pi^T$ finishes training, we use an imitation learning algorithm, DAgger~\cite{ross2011reduction}, to distill $\pi^T$ to a vision-based student policy $\pi^S$ that can access oracle information and let policy help and ease the vision-based policy learning.

\medskip
\noindent
\textbf{State-based teacher policy.} 
In this stage, the inputs are robot state $R_t$, object state $O_t$, the scene point clouds $P_t$, and predicted negative affordance $N_t$. Here the scene point cloud is fused by multi-view depth cameras. Our goal is to learn residual actions $\Delta a_t = \pi^T(S^T_t)$ with predicted negative affordance by PPO~\cite{schulman2017proximal}. The final action is computed with an element-wise addition:
\begin{align}
    a_t = \pi^H(S^T_t) + \pi^T(S^T_t).
\end{align}

\medskip
\noindent
\textbf{Reward function.}
The reward function $r^T$ is defined as:
\begin{align}
    r^T = -r_d^T - r_g^T + r_s^T - r_n^T
\end{align}
where the grasp reward $r_d^T$ penalizes the distance between the dexterous hand and the object, encouraging the hand to maintain contact with the object surface for a secure grasp. The goal reward $r_g^T$ penalizes the distance between the object and the target goal, and the success reward $r_s^T$ provides a bonus when the object successfully reaches the goal. Also the negative affordance reward $r_n^T$ penalizes the dexterous hand to approach the predicted negative affordance. The formal definitions of all rewards are available in our Supp. Mat.


\medskip
\noindent
\textbf{Vision-based student policy.}
For vision-based policy, we only allow it to access information available in the real world, including robot state $R_t$, the scene points clouds $P_t$, and predicted negative affordance $N_t$. 
Then, we distill the teacher policy $\pi^T$ into vision-based student policy $\pi^S$ using DAgger~\cite{ross2011reduction}, i.e., 
\begin{align}
    \pi^S = \operatorname*{arg\,min}_{\pi^S} \| \pi^T(S^T_t) - \pi^S(S^S_t) \|,
\end{align}
where the state for the teacher policy $S^T_t = \{ R_t, O_t, P_t, N_t \}$, and the state for the student policy $S^S_t = \{ R_t, P_t, N_t \}$.

\section{Experiments}
\subsection{Datasets}
\medskip
\noindent
\textbf{UniDexGrasp~\cite{xu2023unidexgrasp}.} This dataset contains 3165 different object instances spanning 133 categories. Evaluation is conducted on these 3,200 seen objects, as well as on 140 unseen objects from seen categories and 100 unseen objects from unseen categories. Each environment is randomly initialized with one object and its initial pose, and the environment consists of a
panoramic 3D point cloud $P_t$ captured from the fixed cameras for vision-based policy learning.

\medskip
\noindent
\textbf{OakInk2~\cite{zhan2024oakink2}.} This dataset record the manipularion processes with pose and shape of the human upper-body and objects. We pre-train our $\pi^H$ using about 2,200 right hand manipulation sequences in this dataset. Also we employ objects in OakInk2 to evaluate the generalization capabilities for grasping.

\begin{table*}[t]
 \centering
 \small
 \setlength{\tabcolsep}{2pt}
 \caption{\textbf{Quantitative comparisons} on UniDexGrasp~\cite{xu2023unidexgrasp} and OakInk2~\cite{zhan2024oakink2}. $HLS$ denotes Human-likeness Score, while $AS$ is Affordance Score.}
   \vspace{-8pt}
 \begin{tabular}{lcccccccccccc}
 \toprule
 \textbf{Method} & 
 \multicolumn{3}{c}{\textbf{Seen Obj.}} & 
 \multicolumn{3}{c}{\textbf{\begin{tabular}{@{}c@{}}Unseen Obj.\\Seen Cat.\end{tabular}}} & 
 \multicolumn{3}{c}{\textbf{\begin{tabular}{@{}c@{}}Unseen Obj.\\Unseen Cat.\end{tabular}}} & 
 \multicolumn{3}{c}{\textbf{OakInk2}} \\ 
 \cmidrule(lr){2-4} \cmidrule(lr){5-7} \cmidrule(lr){8-10} \cmidrule(l){11-13}
 & $Succ\uparrow$ & $HLS\uparrow$ & $AS\downarrow$  & $Succ\uparrow$ & $HLS\uparrow$ & $AS\downarrow$ & $Succ\uparrow$ & $HLS\uparrow$ & $AS\downarrow$  & $Succ\uparrow$ & $HLS\uparrow$ & $AS\downarrow$   \\ 
 \hline
 \multicolumn{13}{c}{\textbf{State-Based Setting}} \\  
 \rowcolor{gray!10}PPO~\cite{schulman2017proximal} & 24.3 & - & - & 20.9 & - & - & 17.2 & - & - & - & - & - \\
 DAPG~\cite{rajeswaran2017learning} & 20.8 & - & - & 15.3 & - & - & 11.1 & - & - & - & - & - \\
 \rowcolor{gray!10}GSL~\cite{jia2022improving} & 57.3 & - & - & 54.1 & - & - & 50.9 & - & - & - & - & - \\
 ILAD~\cite{wu2023learning} & 31.9 & - & - & 26.4 & - & - & 23.1 & - & - & - & - & - \\
 \rowcolor{gray!10}UniDexGrasp~\cite{xu2023unidexgrasp} & 79.4 & 6.9 & 12 & 74.3 & 6.4 & 15 & 70.8 & 6.3 & 18 & 68.4 & 5.9 & 18 \\
 UniDexGrasp++~\cite{wan2023unidexgrasp++} & 87.9 & 5.4 & 28 & 84.3 & 5.2 & 26 & 83.1 & 5.0 & 27 & 79.6 & 4.9 & 28 \\
 \rowcolor{gray!10}DexGrasp Anything~\cite{zhong2025dexgrasp} & 71.2 & - & 20 & 69.1 & - & 18 & 67.3 & - & 22 & 65.9 & - & 24 \\
 \rowcolor[HTML]{D7F6FF} \textbf{\our} & \textbf{89.2} & \textbf{8.6} & \textbf{4} & \textbf{87.7} & \textbf{8.5} & \textbf{7} & \textbf{85.2} & \textbf{8.1} & \textbf{9} & \textbf{82.2} & \textbf{8.2} & \textbf{10} \\
\hline \hline

\multicolumn{13}{c}{\textbf{Vision-Based Setting}} \\  
 \rowcolor{gray!10}PPO~\cite{schulman2017proximal} & 20.6 & - & - & 17.2 & - & - & 15.0 & - & - & - & - & - \\
 DAPG~\cite{rajeswaran2017learning} & 20.8 & - & - & 15.3 & - & - & 11.1 & - & - & - & - & - \\
 \rowcolor{gray!10}GSL~\cite{jia2022improving} & 54.1 & - & - & 50.2 & - & - & 44.8 & - & - & - & - & - \\
 ILAD~\cite{wu2023learning} & 27.6 & - & - & 23.2 & - & - & 20.0 & - & - & - & - & - \\
 \rowcolor{gray!10}UniDexGrasp~\cite{xu2023unidexgrasp} & 73.7 & 6.2 & 16 & 68.6 & 6.1 & 18 & 65.1 & 6.0 & 17 & 62.8 & 5.6 & 20 \\
 UniDexGrasp++~\cite{wan2023unidexgrasp++} & 85.4 & 5.4 & 29 & 79.6 & 5.1 & 25 & 76.7 & 4.8 & 28 & 74.4 & 4.7 & 29 \\
 \rowcolor[HTML]{D7F6FF} \textbf{\our} & \textbf{87.0} & \textbf{8.3} & \textbf{10} & \textbf{82.8} & \textbf{7.8} & \textbf{14} & \textbf{79.2} & \textbf{8.0} & \textbf{15} & \textbf{77.3} & \textbf{7.8} & \textbf{13} \\
 \end{tabular}
 \label{tab:exp}
\end{table*}


\begin{table}[t]
 \centering
 \setlength{\tabcolsep}{7pt}
 \small
 \caption{\textbf{Ablation Study} on UniDexGrasp~\cite{xu2023unidexgrasp} in Seen Object.}
   \vspace{-8pt}
  \begin{tabular}{ccc|ccc}
 \toprule
 HTI & NAA & Distillation & $Succ\uparrow$ & $HLS\uparrow$ & $AS\downarrow$\\ 
 \hline
  \multicolumn{6}{c}{\textbf{State-Based Setting}} \\  
 \rowcolor{gray!10} &  &  & 85.4 & 5.2 & 27 \\
  \checkmark &  &  & 87.9 & 8.2 & 22 \\
  \rowcolor{gray!10}\checkmark & \checkmark &  & \textbf{89.2} & \textbf{8.6} & \textbf{4} \\
\hline \hline
 \multicolumn{6}{c}{\textbf{Vision-Based Setting}} \\  
 \rowcolor{gray!10} &  &  & 70.1 & 5.0 & 27 \\
  &  & \checkmark & 84.9 & 5.6 & 28 \\
  \rowcolor{gray!10} & \checkmark & \checkmark & 85.8 & 7.2 & 13 \\
  \checkmark &  & \checkmark & 86.9 & 8.1 & 20 \\
  \rowcolor{gray!10} \checkmark & \checkmark & \checkmark & \textbf{87.0} & \textbf{8.3} & \textbf{10} \\
 \end{tabular}
 \label{tab:ablation}
\end{table}


\begin{table}[t]
 \centering
 \setlength{\tabcolsep}{3pt}
 \small
 \caption{Results on UniDexGrasp~\cite{xu2023unidexgrasp} in Seen Object in state-based setting.}
   \vspace{-8pt}
  \begin{tabular}{cccc}
 \toprule
 Method & $Succ\uparrow$ & $HLS\uparrow$ & $AS\downarrow$\\ 
 \hline
  \rowcolor{gray!10}UniDexGrasp++~\cite{wan2023unidexgrasp++}  & 87.9 & 5.4 & 28 \\
  UniDexGrasp++ + HTI  & 88.2 & 7.8 & 23 \\
  \rowcolor{gray!10}UniDexGrasp++ + NAA & 88.0 & 5.9 & 19 \\
  UniDexGrasp++ + HTI + NAA  & \textbf{88.8} & \textbf{8.0} & \textbf{12} \\
 \end{tabular}
 \label{tab:more_exp}
\end{table}

\subsection{Metrics}
Following previous works~\cite{xu2023unidexgrasp,wan2023unidexgrasp++,wang2025unigrasptransformer}, each object is randomly rotated and dropped onto the table to enhance the diversity of its initial poses. We report the \textbf{success rate of grasp $Succ$}, \textbf{Human-likeness Score $HLS$}, and \textbf{Affordance Score $AS$} across all objects and grasp attempts. A grasp is considered successful if the object reaches the target goal within 200 steps in simulator. The Human-likeness Score $HLS$ assesses the anthropomorphic quality of the grasp, which is obtained by prompting the Gemini 2.5 Pro~\cite{comanici2025gemini} to analyze a visual sequence of the grasp execution. This metric is specifically to rate the resemblance of the dexterous hand's motion to that of a typical human, yielding a quantitative measure of naturalness. The Affordance Score $AS$, in contrast, evaluates the functional correctness of the grasp by penalizing contact with inappropriate object parts. This metric is calculated using a point cloud of 100 ``negative affordance'' points sampled from our NAA. Specifically, the score is incremented by one for each fingertip that maintains a distance greater than 2cm from any point in this negative set, thus rewarding functionally sound grasps.


\subsection{Implementation Details}
We conduct our experiments in IssacGym~\cite{makoviychuk2021isaac} simulator. During training, 4096 environments are simulated in parallel on an NVIDIA RTX 4090 GPU. For the network architecture, we use MLP with 4 hidden layers (1024,1024,512,512) for the policy network and value network in the state-based setting, and an additional PointNet+Transformer~\cite{mu2021maniskill} to encode the 3D scene point cloud input in the vision-based setting. Other detailed hyperparameters are shown in our Supp. Mat.

\medskip
\noindent
\textbf{Dexterous hand configuration.} We use the Shadow Hand, which features 24 active degrees of freedom (DOFs). The wrist has 6 DOFs controlled by force and torque, while the fingers have 18 active DOFs controlled by joint angles. Specifically, the thumb has 5 DOFs, the little finger has 4, and the remaining three fingers each have 3. Additionally, each finger, excluding the thumb, includes a passive, non-controlled DOF.

\subsection{Comparison with SOTA Methods}
We evaluate \our by first training a state-based policy and then distilling it into a vision-based one. For comparison, we compare \our with several state-of-the-art methods. These include RL and imitation learning algorithms like PPO~\cite{schulman2017proximal}, DAPG~\cite{rajeswaran2017learning}, and ILAD~\cite{wu2023learning}. We also compare against methods with advanced learning paradigms such as GSL~\cite{jia2022improving} (generalist-specialist), UniDexGrasp++~\cite{wan2023unidexgrasp++} (geometry-aware curriculum learning), UniDexGrasp++~\cite{wan2023unidexgrasp++} which proposes a geometry-aware curriculumn learning and generalist-specialist learning, and DexGrasp Anything~\cite{zhong2025dexgrasp}, diffusion-based dexterous grasp generation models. 

Since DexGrasp Anything~\cite{zhong2025dexgrasp} generates only the final static grasp pose rather than a grasping motion, the $HLS$ is not applicable in this setting. To evaluate grasp robustness, we apply a random external force ranging from 0 to 200N to the object to simulate object's gravity.

Tab.~\ref{tab:exp} compares our \our with these SOTA methods using a universal model for dexterous robotic grasping across both state-based, vision-based and even across dataset settings. Our method achieves highest scores in grasping success rate, outperforming other state-of-the-art methods. This improvement stems from our proposed Human Hand Trajectory Imitating, where the policy learns to generate correct and stable grasp poses from human hand motion. This not only greatly enhances the grasp success rate but also leads to a significant improvement in our method's human-likeness score ($HLS$). The significantly lower Affordance Score $AS$ validates the effectiveness of our Negative Affordance-aware Segmentation module. This result indicates that the module successfully guides the policy away from functionally inappropriate regions, leading to grasps at the most suitable locations.

As illustrated in Fig.~\ref{fig:exp}, our method generates a diverse set of grasps. Crucially, it consistently identifies functionally appropriate grasp locations and forms natural hand postures. This combination of functional awareness and naturalness makes the generated poses highly effective for direct application in downstream manipulation tasks.

\begin{figure}
  \centering
  \includegraphics[width=0.95\linewidth]{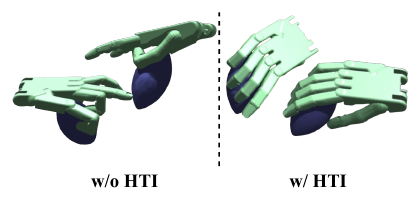}
  \vspace{-7mm}
  \caption{\textbf{Ablation Study} on Human Hand Trajectory Imitating (HTI). Without the human motion prior, the policy converges to a solution that, while potentially successful, is kinematically awkward and non-humanlike.}
  \label{fig:imit}
\end{figure}

\begin{figure}[t]
  \centering
  \includegraphics[width=0.95\linewidth]{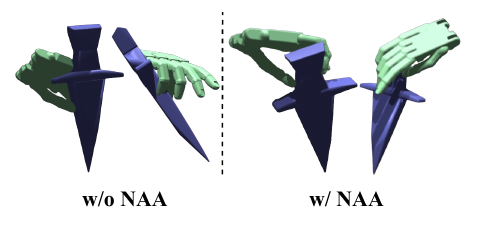}
  \vspace{-7mm}
  \caption{\textbf{Ablation Study} on proposed NAA, which guides the policy to a correct and safte position. The higher Affordance Score ($AS$) for the NAA-guided grasp confirms its superior functional quality.}
  \vspace{-4mm}
  \label{fig:naa}
\end{figure}

\begin{figure}[t]
  \centering
  \includegraphics[width=0.9\linewidth]{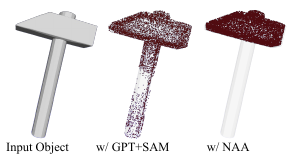}
    \vspace{-10pt}
  \caption{\textbf{Ablation Study} on proposed NAA, which has capability to segment fine-grained negative affordance.}
  \label{fig:naa_seg}
  \vspace{-4mm}
\end{figure}

\subsection{Ablation study}
Unless otherwise specified, the ablation studies are conducted on the seen objects under the state-based setting.

\medskip
\noindent
\textbf{Human Hand Trajectory Imitating.} 
The results in Tab.~\ref{tab:ablation} and Fig.~\ref{fig:imit} show the critical role of pre-training on human trajectories (HTI). When this imitation stage is omitted, the policy, while still capable of finding geometrically stable grasps, produces motions that are kinematically unnatural. This is quantitatively reflected in a sharp increase in the Human-Likeness Score (HLS). Such configurations are not merely an aesthetic issue, they can be inefficient, unpredictable, and detrimental to downstream tasks that require fluid, human-centric interaction.

\medskip
\noindent
\textbf{NAA.} 
As shown in Fig.~\ref{fig:naa_seg}, a naive approach combining an MLLM~\cite{achiam2023gpt} with SAM~\cite{kirillov2023segment} (denoted as GPT+SAM) proves ineffective for this task. This baseline first uses the MLLM's coarse localization ability to provide prompts to SAM. However, because MLLMs like GPT-4V~\cite{achiam2023gpt} excel at image-level understanding but struggle with the fine-grained spatial localization required for segmentation, this process often results in the segmentation of the entire object. In contrast, our NAA module solves this by converting the segmentation task into a simpler classification problem. By first using SAM to generate accurate mask proposals and then using CLIP to select the one with the highest semantic similarity to the negative affordance description, NAA achieves precise segmentation.

As shown in Fig.~\ref{fig:naa} and Tab.~\ref{tab:ablation}, the guidance from NAA results in a significant decrease in the Affordance Score $AS$, which indicates that the policy successfully learns to make contact at more rational and safer locations on the object. By generating functionally sound grasps, \our greatly improves the feasibility of performing downstream tasks.

\medskip
\noindent
\textbf{Teacher-student distillion.} 
Without the teacher-student distillion (Distillation) grasping accuracy
decreases significantly. This is primarily due to the lack of privileged information guidance, which makes it challenging for the single-stage RL policy to learn the position to grasp. As shown in Tab.~\ref{tab:ablation}, the policy without teacher-student distillation demonstrates lower grasping success rate.

\subsection{Extension to Other Grasping Methods}
Notably, our proposed modules demonstrate strong generalizability by significantly enhancing other RL-based methods, such as UniDexGrasp++~\cite{wan2023unidexgrasp++}. Specifically, the Human Hand Trajectory Imitating (HTI) module markedly improves the naturalness and human-likeness of its generated poses. Simultaneously, the Affordance-aware Residual Learning, guided by negative affordances from our Negative Affordance-aware Segmentation (NAA), substantially boosts the semantic appropriateness of its grasp locations on the object, as shown in Tab.~\ref{tab:more_exp}.

\section{Conclusion}

\medskip
\noindent

In this paper, we present \our, a novel framework for generating dexterous grasps that are not only successful but also human-like and functionally correct. 
Our key insight is that the challenges of naturalness and functional correctness can be effectively decoupled and then synergized: a strong motion prior learned from human data constrains the policy to a manifold of natural poses, while a visual understanding of negative affordances guides the policy to safe and appropriate contact regions. 
Extensive experiments validate that this approach significantly outperforms state-of-the-art baselines in success rate, pose naturalness, and contact appropriateness. We believe this work lays a crucial foundation for more general-purpose embodied agents and opens new avenues for research in dexterous manipulation.

\textbf{Limitation.} 
A limitation of our approach stems from its reliance on a fixed set of six rendered views for negative affordance prediction, which can fail to capture all functionally relevant parts on geometrically complex or concave objects. This can lead to imprecise negative affordance segmentation due to occlusion. Future work could overcome this by adopting volumetric-based affordance learning on implicit 3D representations, which are inherently robust to viewpoint-specific occlusions.

\section{Acknowledgement}
This work was supported by Central Guidance for Local Science and Technology Development Fund (ZYYD2025QY19), an Bingtuan Science and Technology Program (2022DB005). This work was supported by DAMO Academy through Academy Research Intern Program.

\bibliography{aaai2026}

\clearpage

\section{Details about Our Method}

\begin{algorithm}
\caption{Overall AffordDex Framework}
\label{alg:main_en_revised}
\begin{algorithmic}[1] 
\Require Human hand motion dataset $\mathcal{D}_H$, object dataset $\mathcal{D}_O$
\Ensure Final vision-based grasping policy $\pi^S$

\State {\textbf{Stage 1: Human Hand Trajectory Imitating}}
\State Pre-train a base policy $\pi^H$ on $\mathcal{D}_H$ via imitation learning to establish a human motion prior.

\State {\textbf{Stage 2: Affordance-aware Residual Learning and Distillation}}
\State Generate negative affordance point clouds $\{n_O\}_{O \in \mathcal{D}_O}$ for all objects using the NAA module.
\State Learn a state-based teacher policy $\pi^T$ by fine-tuning $\pi^H$ with a residual module, guided by the negative affordances $\{n_O\}$.
\State Distill the teacher policy $\pi^T$ into a vision-based student policy $\pi^S$ via a teacher-student framework (e.g., DAgger).

\State \textbf{return} Final vision-based policy $\pi^S$.
\end{algorithmic}
\end{algorithm}

\section{Details about Baselines}

\medskip
\noindent
\textbf{PPO}. We use Proximal Policy Optimization (PPO)~\cite{schulman2017proximal}, a standard on-policy reinforcement learning (RL) algorithm, as the foundation for our RL-based training stages.

\medskip
\noindent
\textbf{DAPG}. To leverage expert data, we employ Demo-Augmented Policy Gradient (DAPG)~\cite{rajeswaran2017learning}. This imitation learning (IL) method accelerates policy training by combining a policy gradient loss with a behavior cloning loss on expert demonstrations. These demonstrations are generated via motion planning.

\medskip
\noindent
\textbf{GSL}. We also include Generalist-Specialist Learning (GSL)~\cite{jia2022improving}, a three-stage training paradigm. It first trains a generalist policy, then fine-tunes specialized policies on difficult task subsets, and finally uses demonstrations from these specialists to train a final, more capable generalist via IL. For fair comparison, our GSL implementation uses PPO for the RL components and DAPG for the final IL stage.

\medskip
\noindent
\textbf{ILAD}. To further improve generalization, we compare against ILAD~\cite{wu2023learning}, an IL method that builds upon DAPG. It introduces an auxiliary objective that forces the policy to learn a geometric object representation from the same motion-planned demonstrations, enhancing its adaptability.

\medskip
\noindent
\textbf{UniDexGrasp and UniDexGrasp++}. 
UniDexGrasp~\cite{xu2023unidexgrasp} is a two-stage learning method for dexterous grasping. In the first stage, it trains a state-based teacher policy using Reinforcement Learning. To manage training across numerous objects, it introduces Object Curriculum Learning (OCL), which starts with a single object and gradually incorporates more objects from similar semantic categories. In the second stage, this proficient teacher policy is distilled into a vision-based student policy using DAgger~\cite{ross2011reduction}, enabling it to operate from point cloud inputs.

Building upon this foundation, \textbf{UniDexGrasp++} aims to significantly enhance the policy's generalizability across thousands of object instances with diverse geometries. It argues that curriculum based on semantic categories can be suboptimal and instead proposes two novel, geometry-centric techniques: Geometry-aware Curriculum Learning (GeoCurriculum) and Geometry-aware iterative Generalist-Specialist Learning (GiGSL). GeoCurriculum organizes the training progression based on object geometric features rather than categories, creating a more effective learning path for grasping. GiGSL further refines the policy by iteratively training a generalist model on all objects and specialist models on geometrically challenging subsets. These innovations lead to a more robust universal grasping policy that substantially outperforms its predecessor.

\medskip
\noindent
\textbf{DexGrasp Anything}.
DexGrasp Anything~\cite{zhong2025dexgrasp} is a recent method for generating high-quality, static dexterous grasp poses. It utilizes a diffusion-based generative model to address object diversity and hand complexity. Its core innovation is the direct integration of physical constraints (e.g., collision-freeness, stability) into both the training and sampling phases of the diffusion process. Since it generates only the final static grasp pose rather than a grasping motion, we apply a random external force ranging from 0 to 200N to the object to simulate object's gravity.

For PPO, DAPG, GSL, ILAD, UniDexGrasp, and UniDexGrasp++, we maintain the same experimental settings as reported in the UniDexGrasp++ paper~\cite{wan2023unidexgrasp++} to ensure a fair comparison. For the DexGrasp Anything baseline, we directly utilized the officially released, pretrained model weights provided by the authors.



\section{Experiment Details}
We use PPO to train Reinforcement learning. We use DAgger-based policy distillation to distill the state-based policy into a vision-based one. We set $w_{f}=1$ in Eq.~\ref{eq:eq1}. We follow FGVP and set NMS IoU threshold in NAA to 0.7 (robust in practice).

For UniDexGrasp in Seen Object, we conduct 5 independent runs using fixed random seed of 42, to assess the performance consistency, reported in the table below.


\begin{figure}
  \centering
  \includegraphics[width=\linewidth]{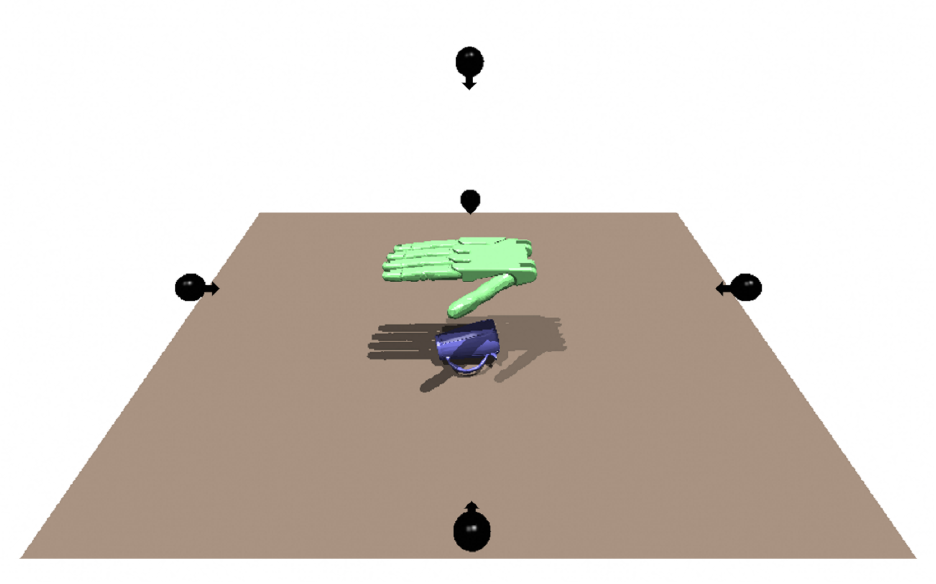}
  \vspace{-7mm}
  \caption{Illustration of the simulation environment.}
  \label{fig:imit}
\end{figure}

\subsection{Experiment Setup}

\medskip
\noindent
\textbf{State Definition}.
The full state of the teacher policy $S^T_t = \{ R_t, O_t, P_t, N_t \}$. The complete object point clouds are assumed to be perfectly accurate. Object states $O_t$, including positions, rotations, and velocities, are directly accessible. To accelerate the training process, we sample 1024 points from the object and the hand in the scene point cloud $P_t$.

The full state of the student policy $S^S_t = \{ R_t, P_t, N_t \}$. Partial object point clouds are reconstructed and segmented from depth data captured by five cameras around the table. The hand-object distance is computed using the partial object point cloud in the vision-based setting.

\medskip
\noindent
\textbf{Action Space}.
The action space consists of the motor commands for the 24 actuators of the dexterous hand. The first 6 actuators control the global position and orientation, while the remaining 18 control the finger joints. We normalize the action range to $(-1, 1)$.

\medskip
\noindent
\textbf{Camera Setup}.
Following a similar approach to UniDexGrasp++~\cite{wan2023unidexgrasp++}, five RGBD cameras are mounted around the table, as shown in Fig.~\ref{fig:imit}. The cameras are positioned relative to the table center at coordinates $(0.0, 0.0, 0.55)$, $(0.5, 0.0, 0.15)$, $(-0.5, 0.0, 0.15)$, $(0.0, 0.5, 0.15)$, $(0.0, -0.5, 0.15)$, with their focal points aligned at $(0, 0, 0.15)$. In the vision-based setting, the depth images captured by these cameras are fused to generate a scene point cloud, from which the partial point cloud of the object is segmented.

\begin{table}[t]
 \centering
 \setlength{\tabcolsep}{9pt}
 \small
 \caption{\textbf{Ablation Study} on UniDexGrasp dataset~\cite{xu2023unidexgrasp} in Seen Object in state-based setting.}
  \begin{tabular}{c|ccc}
 \toprule
 Configuration & $Succ\uparrow$ & $HLS\uparrow$ & $AS\downarrow$\\ 
 \hline
  \rowcolor{gray!10}$\lambda^{smooth}=0.02$ & 89.0 & 7.9 & \textbf{4} \\
  $\lambda^{smooth}=0.05$ & \textbf{89.2} & \textbf{8.6} & \textbf{4} \\
  \rowcolor{gray!10}$\lambda^{smooth}=0.1$ & 88.2 & 8.5 & 6 \\
  \hline
  $\lambda^{finger}=0.5$ & 87.8 & 8.5 & \textbf{4} \\
  \rowcolor{gray!10}$\lambda^{finger}=0.8$ & \textbf{89.2} & \textbf{8.6} & \textbf{4} \\
  $\lambda^{finger}=1.0$ & 88.5 & 8.2 & 6 \\
 \end{tabular}
 \label{tab:ablation_parms-1}
\end{table}


\medskip
\noindent
\textbf{Reward Function for Human Hand Trajectory Imitating}.
The goal of our human hand trajectory imitation reward, $r^H$ is to encourage the agent to mirror the articulation and motion smoothness of a reference human trajectory. It is formulated as a weighted sum of two key components: a finger imitation reward and a smoothness reward.

The finger imitation reward $r^H_{\text{finger}}$ encourages the agent's hand to accurately track the reference finger poses. We define this reward based on the squared Euclidean distance between corresponding keypoints on the agent's hand and the human reference hand:
\begin{align}
    r^H_{\text{finger}} = \sum_{f=1}^{F} w_{f} \cdot \exp \left( -\lambda_{f} \left\| \mathbf{j}_{d, f} - \mathbf{j}_{h, f} \right\|_{2}^{2} \right),
\end{align}
where $\mathbf{j}_{d, f}$ is the position of the $f$-th keypoint on the dexterous hand, and $\mathbf{j}_{h, f}$ is its corresponding target position from the reference trajectory. The parameters $w_{f}$ and $\lambda_{f}$ are set according to the anatomical group to which keypoint $f$ belongs. Specifically, we group keypoints into two levels:
\begin{itemize}
    \item For keypoints $f$ in \textbf{Level-1} (base joints), we use a stricter decay rate of $\lambda_f = 50$.
    \item For keypoints $f$ in \textbf{Level-2} (middle joints), we use $\lambda_f = 40$.
\end{itemize}
The weights $w_f$ are configured to combine these rewards appropriately. This hierarchical parameterization allows the model to prioritize accurate positioning of the more critical base joints.

The smoothness reward $r^H_{\text{smooth}}$ is designed to alleviate jerky motions, penalizing the power exerted on each joint, defined as the element-wise product of joint velocities and torques
\begin{align}
    r_{t}^{\text{smooth}} = -w_{\text{smooth}} \sum_{i=1}^{n} \left| \tau_{t,i} \cdot \dot{q}_{t,i} \right|,
\end{align}
where $w_{\text{smooth}}$ is a positive weighting coefficient, $n$ is the total number of actuated joints, and $\tau_{t,i}$ and $\dot{q}_{t,i}$ are the torque and angular velocity of the $i$-th joint at timestep $t$, respectively. The product $\tau_{t,i} \cdot \dot{q}_{t,i}$ represents the instantaneous power exerted by the actuator of joint $i$. By taking the absolute value, we penalize any high-power action, including both strong acceleration and aggressive braking, thus encouraging the policy to learn energy-efficient and hardware-friendly motions.

The final reward in this stage is:
\begin{align}
    r^H = \lambda^{\text{smooth}} r_{t}^{\text{smooth}} + \lambda^{\text{finger}} r^H_{\text{finger}},
\end{align}
where we set $\lambda^{\text{smooth}}=0.05$ and $\lambda^{\text{finger}}=0.8$. We also conduct ablation studies on these weights in Tab.~\ref{tab:ablation_parms-1}.


\medskip
\noindent
\textbf{Reward Function for Affordance-aware Residual Learning}.
The reward function described in Eq.(4) of the main paper comprises four components: a grasp reward $r_d^T$, a goal-reaching reward $r_g^T$, a success bonus $r_s^T$, and a negative affordance penalty $r_n^T$. Each component is detailed below.

The grasp reward $r_d^T$ encourages the hand to approach and stay close to the object. It is defined as a penalty proportional to the distance between the hand and the object's center:
\begin{align}
    r_{d}^T = \lambda_d^T \| p_{\text{dex}} - p_{\text{obj}} \|,
\end{align}
where $p_{\text{dex}}$ and $p_{obj}$ are the Cartesian position of the dexterous hand and the object, respectively. We set $\lambda_d^T$ to -1.

The goal reward $r_g^T$ guides the object towards the target goal. It is structured as a penalty on the distance between the object and the goal:
\begin{align}
    r_{g}^T = \lambda_g^T  \| p_{\text{obj}} - p_{\text{goal}} \|,
\end{align}
where $p_{\text{goal}}$ is the position of the target goal. We set $\lambda_g^T$ to -1.

A bonus is provided upon task completion. This is triggered when the object enters a small threshold radius around the goal:
\begin{align}
r_{s}^T = \lambda_s^T \mathbb{I}(\| p_{\text{obj}} - p_{\text{goal}} \| < \alpha_s),
\end{align}
where $\mathbb{I}(\cdot)$ is the indicator function, which evaluates to 1 if the condition is true and 0 otherwise. $\alpha_s$ is set to 0.05. We set $\lambda_s^T$ to 1.

The negative affordance reward $r_n^T$ penalizes the dexterous hand fingertips to approach the predicted negative affordance:
\begin{align}
    r_{n}^T = \lambda_n^T \sum_{f \in \mathcal{K}_c} \mathbb{I} \left( \min_{p_n \in \mathcal{P}_n} \| p_{\text{tip}}^f - p_n \| < \alpha_n \right),
\end{align}
where $\alpha_n$ is set to 0.03, $\mathcal{K}_c$ is the set of the hand's fingertips, $p_{\text{tip}}^f$ is the position of fingertip $f$, and $\mathcal{P}_n$ is the set of points representing the negative affordance. We set $\lambda_n^T = -10$. \textbf{Ablation studies are conducted on these weights to evaluate their contribution. The results are presented in Hyperparameters part.}

\begin{table}[t]
 \centering
 \setlength{\tabcolsep}{9pt}
 \small
 \caption{\textbf{Ablation Study} on UniDexGrasp dataset~\cite{xu2023unidexgrasp} in Seen Object in state-based setting.}
  \begin{tabular}{c|ccc}
 \toprule
 Configuration & $Succ\uparrow$ & $HLS\uparrow$ & $AS\downarrow$\\ 
 \hline
  \rowcolor{gray!10}$\lambda_d^T=-0.5$ & 84.2 & 7.2 & 5 \\
  $\lambda_d^T=-1$ & \textbf{89.2} & \textbf{8.6} & \textbf{4} \\
  \rowcolor{gray!10}$\lambda_d^T=-2$ & 88.4 & 8.4 & \textbf{4} \\
  \hline
  $\lambda_g^T=-0.5$ & 87.9 & 8.1 & \textbf{4} \\
  \rowcolor{gray!10}$\lambda_g^T=-1$ & \textbf{89.2} & \textbf{8.6} & \textbf{4} \\
  $\lambda_g^T=-2$ & 88.9 & 8.3 & 6 \\
  \hline
  \rowcolor{gray!10}$\lambda_s^T=0.5$ & 83.1 & 7.0 & 6 \\
  $\lambda_s^T=1$ & \textbf{89.2} & \textbf{8.6} & \textbf{4} \\
  \rowcolor{gray!10}$\lambda_s^T=2$ & 89.0 & 8.5 & 5 \\
  \hline
  $\lambda_n^T=-5$ & 89.1 & 8.5 & 16 \\
  \rowcolor{gray!10}$\lambda_n^T=-10$ & \textbf{89.2} & \textbf{8.6} & 4 \\
  $\lambda_n^T=-20$ & 87.1 & 8.0 & \textbf{0} \\
 \end{tabular}
 \label{tab:ablation_parms}
\end{table}


\medskip
\noindent
\textbf{Human-likeness Score ($HLS$)}.
The Human-likeness Score (HLS) is designed to quantify the anthropomorphic quality of the generated grasping motion. We leverage the advanced multi-modal capabilities of Gemini 2.5 Pro~\cite{comanici2025gemini} to serve as an expert evaluator.

For each grasp, a video sequence of the complete grasp execution is provided as input to the model. The model is then prompted with a specific set of instructions to assess the motion based on key kinematic criteria. The exact prompt provided to the model is as follows:

\begin{quote}
    \textbf{Expert Role:} You are an expert in hand kinematics evaluation.
    
    \vspace{0.5em}
    
    \textbf{Task:} Evaluate the similarity between the robotic hand motion in the simulation video and that of a real human hand, based on the following three criteria:
    \begin{itemize}
        \item Motion trajectory
        \item Velocity smoothness
        \item Joint coordination
    \end{itemize}
    
    \vspace{0.5em}
    
    \textbf{Output Format:} Return only a valid JSON, following this format: \\
    \texttt{\{ "score": <1-10> \}}.
\end{quote}


\medskip
\noindent
\textbf{VLM-based Negative Affordance Identification}
VLMs struggle to interpret non-textured 3D meshes, as these models primarily rely on rich visual cues learned from images. To bridge this gap, we first apply procedural texturing to the raw meshes using Tex-Painter~\cite{zhang2024texpainter}, which generates semantically plausible textures based on geometric analysis. Next, we render the textured object from six cardinal directions to create a multi-view image set, $I$, as a holistic visual representation. While this may not capture all concavities in highly complex objects, we found it provides a sufficient basis for affordance prediction for the objects in our benchmark datasets.

We then query GPT-4V~\cite{achiam2023gpt} with the multi-view images to elicit a description of the object's negative affordances. The exact prompt is as follows:

\begin{quote}
    \textbf{Expert Role:} You are an expert in identifying which part of a physical object should not be touched, especially in robotic grasping tasks.
    
    \vspace{0.5em}
    
    \textbf{Task:} You will be given 6-view images of an object. Your task is:
    \begin{itemize}
        \item Identify one non-touchable part.
        \item Infer the object's identity from the image name.
    \end{itemize}
    
    \vspace{0.5em}
    
    \textbf{Output Format:} Return one sentence only, following this format: \\
    \texttt{"This is a \_\_\_\_\_. \_\_\_\_\_ of \_\_\_\_\_\_ should not be touched."}
\end{quote}


\medskip
\noindent
\textbf{Points selected in Negative Affordance-aware Segmentation}
The core idea is to systematically prompt SAM with a dense, regular grid of points covering the entire image. This process ensures that objects of various sizes and locations are likely to be "hit" by at least one point prompt, triggering SAM to segment them. The procedure consists of the following steps:

\begin{enumerate}
    \item \textbf{Grid Definition:} We define a regular grid of points to be overlaid on the input image. The density of this grid is controlled by a single hyperparameter, $g = 16$, which represents the number of points along each of the image's dimensions. This results in a total of $g^2$ keypoints.

    \item \textbf{Keypoint Generation:} For an input image $I$ with width $W$ and height $H$, the set of grid points $G$ is generated. The coordinates $(x_i, y_j)$ for each point are calculated by uniformly spacing them across the image dimensions. The formula for a point at grid position $(i, j)$ is:
    \begin{equation}
        G = \left\{ (x_i, y_j) \mid x_i = \frac{i}{g+1} \cdot W, \quad y_j = \frac{j}{g+1} \cdot H \right\}
    \end{equation}
    for all $i, j \in \{1, 2, \dots, g\}$. Using $g+1$ as the denominator ensures that the points are placed within the image boundaries and not on the absolute edges.

    \item \textbf{Prompting SAM:} Each of the $g^2$ points in the set $G$ is used as an individual positive point prompt for SAM. This yields a large collection of raw segmentation masks, $\mathcal{M}_{\text{raw}} = \text{SAM}(I, G)$, where many masks may be redundant or highly overlapping.

    \item \textbf{Filtering with NMS:} To produce a refined set of unique proposals, we apply Non-Maximum Suppression (NMS) to the raw masks. The NMS algorithm filters out duplicate masks based on the Intersection over Union (IoU) of their corresponding bounding boxes, retaining only the most confident and distinct object proposals.
\end{enumerate}

\medskip
\noindent
\textbf{Hyperparameters}.
As shown in Tab.~\ref{tab:ablation_parms}, we conduct a series of ablation studies to validate our design choices and analyze the model's sensitivity to key hyperparameters. Our investigation focuses on the weights of our proposed reward function. The goal is to find a robust set of parameters that balances efficient learning with task performance.

The results indicate that our chosen default configuration is robust and well-justified. For the grasp reward weight $\lambda^T_d$ and the goal reward weight $\lambda_g^T$, we observe an optimal value of -1.0.  Deviating from this value, by either decreasing the penalty to -0.5 or increasing it to -2.0, leads to a degradation in both $Succ$ and $HLS$.  This suggests a critical balance: a penalty that is too weak fails to prevent undesirable actions, while an overly strong penalty can cause the policy to focus myopically on avoiding penalties rather than achieving a stable grasp, ultimately degrading task success. Similarly, for the bonus reward weight $\lambda^T_s$, the optimal value is 1.0.  A lower weight of 0.5 fails to sufficiently incentivize the target behavior, resulting in a significant drop in performance. Conversely, increasing the weight to 2.0 offers no additional benefit and slightly hinders performance, indicating that an excessive bonus can also be detrimental. Similarly, the negative affordance weight $\lambda_n^T$ requires careful tuning. An overly strong penalty can make the agent too conservative, causing it to avoid the target object altogether in its effort to steer clear of negative affordances. Conversely, an insufficient penalty fails to effectively deter the agent from frequently approaching these undesirable regions.

\section{Future Experiments}
We chose simulation to enable large-scale generalization experiments. The policy is vision-based and avoids privileged information, designing it for sim-to-real transfer, which is a key future goal.



\end{document}